\documentclass{article}

\usepackage[preprint]{neurips_2026}

\usepackage[utf8]{inputenc}
\usepackage[T1]{fontenc}

\usepackage{hyperref}
\usepackage{url}
\usepackage{booktabs}
\usepackage{amsfonts}
\usepackage{nicefrac}
\usepackage{microtype}
\usepackage{xcolor}

\usepackage{amsmath}
\usepackage{amssymb}
\usepackage{amsthm}
\usepackage{graphicx}
\usepackage{multirow}
\usepackage{array}
\usepackage{algorithm}
\usepackage{algpseudocode}
\usepackage{mathtools}
\usepackage{bm}

\title{
Auditable Climate Risk Intelligence from Fragmented ESG Data:
Deterministic Orchestration and Imbalance-Aware Learning
for Scope 1--3 Validation
}

\author{
Karan Sehgal \\
Kent Business School \\
University of Kent \\
Canterbury, United Kingdom \\
\texttt{K.Sehgal@kent.ac.uk}
\And
Khawar Naveed Bhatti \\
Kent Business School \\
University of Kent \\
Canterbury, United Kingdom \\
\texttt{K.Bhatti@kent.ac.uk}
}

\begin{document}

\maketitle


\begin{abstract}

Corporations pursuing net-zero commitments increasingly face a systemic engineering challenge: ESG and climate-risk data remain fragmented across heterogeneous Scope 1, Scope 2, and Scope 3 reporting environments, while conventional validation pipelines lack mechanisms for provenance-aware auditability, hidden drift detection, and reproducibility-oriented governance.

This paper proposes a deterministic climate-risk intelligence framework integrating single-source-of-truth orchestration, temporal anomaly detection, imbalance-aware ensemble learning, and explainability-oriented governance for auditable ESG validation infrastructure.

To support open reproducibility, we construct and release a synthetic ESG validation benchmark whose disclosure distributions, anomaly prevalence, and missingness structure are calibrated against publicly reported characteristics of the GHG Protocol, PCAF, and ISSB reporting standards. The proposed workflow combines this benchmark with public climate-risk hazard datasets, proxy emissions features, and accounting-aligned data-quality logic within a deterministic event-driven orchestration pipeline designed to improve validation traceability and governance reliability.

The methodology incorporates temporal drift analysis, SMOTE-based rare-event optimization, ensemble learning architectures, provenance-aware orchestration semantics, and TreeSHAP-based interpretability for governance inspection and audit reconstruction.

Comparative experimentation evaluates the framework against statistical classifiers, anomaly-detection methods, temporal forecasting baselines, and a threshold-based validation system using classification metrics (recall, F1, ROC-AUC), calibration metrics (Expected Calibration Error, Brier score), and a governance-oriented \emph{audit trace completeness} metric measuring the fraction of flagged anomalies for which a deterministic source-to-escalation provenance chain can be reconstructed. Results are reported as the mean and standard deviation across stratified five-fold cross-validation, with paired significance testing against the strongest baseline.

The proposed workflow reframes ESG reporting from passive disclosure aggregation toward deterministic climate-risk governance infrastructure supporting reproducibility, explainability, and operational auditability within regulated enterprise environments.

\end{abstract}


\section{Introduction}

Corporations pursuing net-zero commitments increasingly operate under fragmented ESG reporting environments involving heterogeneous Scope 1, Scope 2, and Scope 3 disclosure systems distributed across suppliers, financial systems, sustainability platforms, and geographically diverse reporting infrastructures.

Conventional ESG validation workflows frequently rely upon passive extraction, threshold-based reconciliation procedures, and manually intensive audit processes lacking deterministic orchestration semantics, provenance-aware traceability, and reproducibility-oriented governance engineering.

These limitations become increasingly significant under regulated enterprise environments where climate-risk disclosures influence institutional reporting obligations, financed-emissions accounting, operational risk evaluation, and sustainability-linked financial decision making.

Enterprise climate-risk environments additionally exhibit sparse anomaly distributions and highly imbalanced operational conditions represented as:

\[
P(y = 1) \ll P(y = 0)
\]

where governance-critical anomalies constitute only a small fraction of total reporting observations.

Under such conditions, conventional optimization procedures may produce deceptively strong aggregate performance while exhibiting weak minority-event sensitivity and elevated false-negative behavior under governance-critical edge cases.

Additionally, ESG reporting systems remain vulnerable to:

\begin{itemize}
    \item provenance inconsistencies,
    \item delayed supplier disclosures,
    \item climate-transition drift,
    \item hidden reconciliation conflicts,
    \item and incomplete emissions reporting.
\end{itemize}

This paper proposes a deterministic climate-risk intelligence framework integrating:

\begin{itemize}
    \item deterministic orchestration,
    \item provenance-aware governance,
    \item temporal drift detection,
    \item imbalance-aware ensemble learning,
    \item and explainability-oriented audit infrastructure.
\end{itemize}

The proposed workflow combines heterogeneous ESG reporting environments, climate-risk hazard datasets, temporal anomaly modeling, and governance-oriented explainability mechanisms within a reproducibility-aware orchestration architecture designed to improve auditability and operational reliability under fragmented enterprise disclosure conditions.

The broader objective is to transition ESG reporting from passive disclosure aggregation toward deterministic governance engineering supporting trustworthy climate-risk intelligence infrastructure.

The primary contributions of this work are summarized as follows:

\begin{itemize}
    \item A deterministic orchestration framework for provenance-aware ESG validation under fragmented enterprise reporting environments.

    \item An imbalance-aware climate-risk intelligence workflow integrating SMOTE optimization, ensemble learning, and temporal anomaly detection.

    \item A governance-oriented explainability architecture combining TreeSHAP-based interpretability with audit reconstruction and provenance-aware traceability.

    \item An \emph{audit trace completeness} metric that quantifies deterministic provenance reconstruction for flagged anomalies, making the governance-engineering objective directly measurable.

    \item A reproducibility-oriented experimentation framework incorporating a calibrated synthetic ESG benchmark, climate-risk enrichment, temporal drift modeling, cross-validated evaluation, and governance-sensitive metrics.
\end{itemize}


\begin{figure*}[t]
\centering
\includegraphics[width=\textwidth]{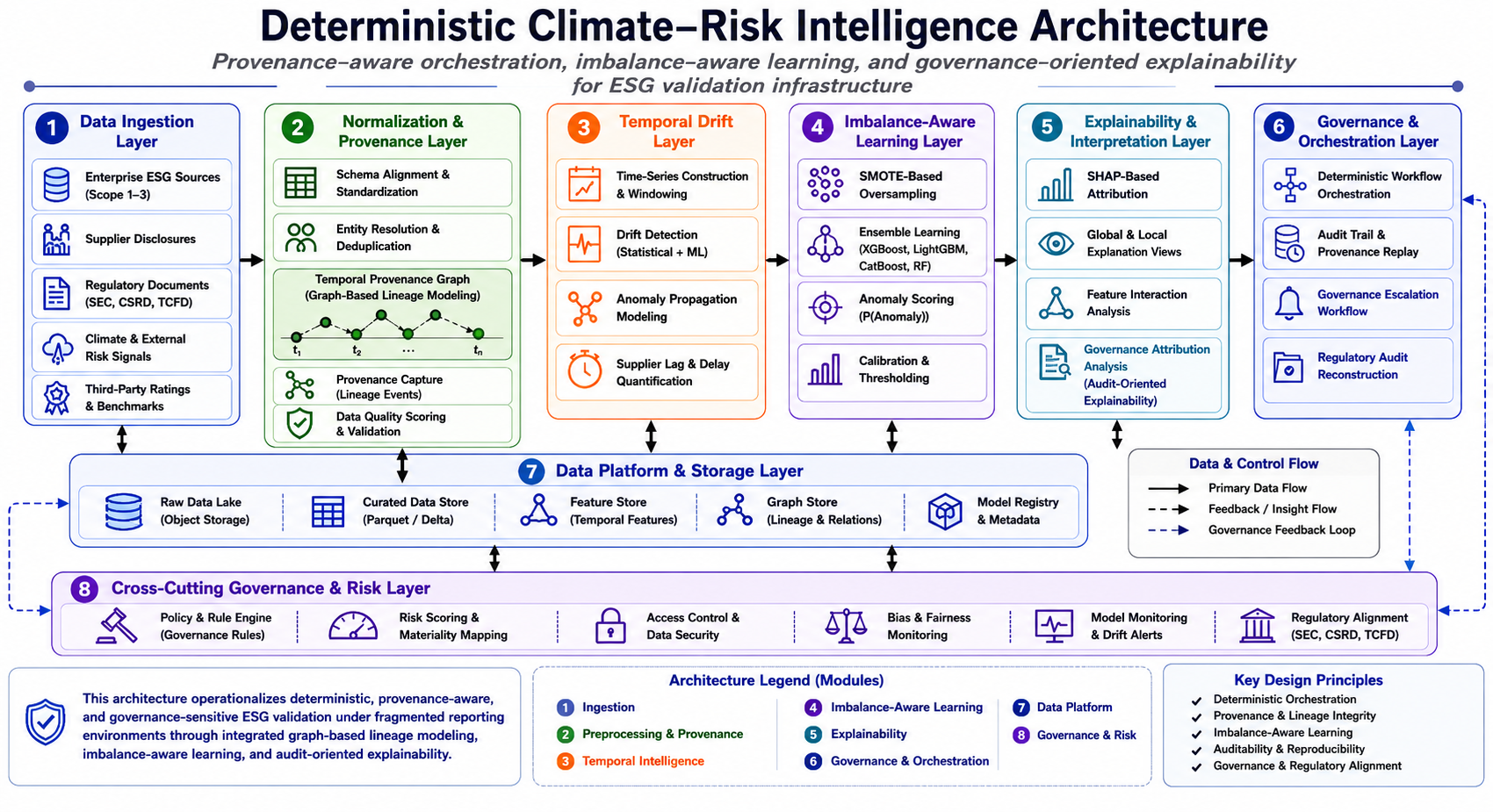}
\caption{
Deterministic climate-risk intelligence architecture integrating provenance-aware orchestration, imbalance-aware learning, temporal provenance graph modeling, and governance-oriented explainability for ESG validation infrastructure.
}
\label{fig:architecture}
\end{figure*}


\section{Related Work}

\subsection{ESG Reporting and Climate-Risk Governance}

Enterprise climate-risk reporting has evolved rapidly under governance frameworks including the GHG Protocol, PCAF, TCFD, ISSB, and CSRD \cite{ghgprotocol2023,pcaf2022,issb2023}.

These frameworks establish accounting-oriented disclosure requirements but provide comparatively limited technical guidance regarding deterministic orchestration, provenance-aware validation, and reproducibility-oriented governance engineering.

Recent climate-risk systems increasingly incorporate NGFS transition scenarios, Copernicus climate infrastructure, WRI Aqueduct hazard datasets, and ThinkHazard environmental-risk intelligence for enterprise stress-testing and climate-exposure modeling \cite{ngfs2023,copernicus2023,aqueduct2023,thinkhazard2023}.

However, existing ESG validation workflows remain heavily dependent upon static reconciliation procedures lacking temporal anomaly awareness, governance-oriented orchestration semantics, and deterministic audit reconstruction.

\subsection{Machine Learning for Climate-Risk Intelligence}

Machine learning architectures including Random Forests \cite{breiman2001random}, Gradient Boosting, XGBoost \cite{chen2016xgboost}, LightGBM \cite{ke2017lightgbm}, and CatBoost \cite{prokhorenkova2018catboost} have demonstrated strong predictive capability under nonlinear enterprise-risk environments.

Recent work additionally explores emissions estimation, climate-risk forecasting, and governance-sensitive anomaly detection using ensemble-learning architectures.

However, governance-critical ESG failures remain comparatively sparse and operationally imbalanced, frequently producing unstable minority-event sensitivity and elevated false-negative behavior under conventional optimization workflows.

Imbalance-aware learning procedures including SMOTE \cite{chawla2002smote} therefore remain important for governance-oriented anomaly detection under fragmented enterprise reporting environments.

\subsection{Concept Drift and Temporal Instability}

Enterprise ESG reporting systems additionally exhibit temporal instability involving delayed disclosures, supplier volatility, evolving sustainability standards, and climate-transition uncertainty.

Prior research on concept drift and adaptive learning highlights the importance of time-aware validation and non-stationary anomaly modeling within continuously evolving enterprise systems \cite{gama2014drift}.

Temporal governance degradation may emerge through:

\begin{itemize}
    \item concept drift,
    \item covariate drift,
    \item label drift,
    \item and reporting-frequency instability.
\end{itemize}

These forms of temporal instability become particularly significant within climate-risk intelligence systems operating under evolving regulatory and environmental conditions.

\subsection{Explainability and Trustworthy AI}

Explainability-oriented methodologies including SHAP \cite{lundberg2017unified} and its tree-ensemble specialization TreeSHAP \cite{lundberg2020treeshap} have emerged as important mechanisms for improving enterprise AI transparency, audit reconstruction, and governance inspection. TreeSHAP is particularly relevant here because it computes exact Shapley values for tree-based models in low-order polynomial time, making attribution tractable across the high-dimensional feature space used in this work.

Recent trustworthy-AI governance frameworks including the NIST AI Risk Management Framework \cite{nist2023airmf}, OECD trustworthy-AI principles \cite{oecdai}, and reproducibility-oriented machine-learning guidance \cite{pineau2021reproducibility} additionally emphasize calibration reliability, operational reproducibility, governance-aware oversight, and audit consistency.

However, explainability alone does not guarantee deterministic orchestration, provenance-aware governance, or reproducibility-oriented audit infrastructure.

This paper therefore positions explainability as one component within a broader deterministic governance-engineering architecture for trustworthy climate-risk intelligence systems.


\section{Dataset and Problem Setup}

\subsection{A Calibrated Synthetic ESG Benchmark}

To enable open, fully reproducible experimentation without disclosing commercially sensitive enterprise records, we construct and release a \emph{synthetic} ESG validation benchmark. The benchmark is generated by a documented data-generating process (DGP) rather than extracted from a single proprietary system; this design choice trades away real-world idiosyncrasy in exchange for full reproducibility, public availability, and controllable anomaly prevalence. Synthetic benchmarking is a standard practice for governance-sensitive domains in which production data cannot be shared.

The DGP proceeds in four stages. First, base disclosure records are sampled across Scope 1, Scope 2, and Scope 3 fields, with marginal distributions and missingness rates calibrated against publicly reported characteristics of the GHG Protocol, PCAF, and ISSB reporting standards \cite{ghgprotocol2023,pcaf2022,issb2023}. Second, provenance metadata, confidence bands, and temporal reporting signals are attached to each record. Third, the six governance failure modes of Section~\ref{sec:failuremodes} are injected at a controlled prevalence to produce labelled governance anomalies. Fourth, records are enriched with public climate-risk indicators (Section~\ref{sec:publicclimate}). The anomaly label is therefore generated jointly with the failure-injection process, giving an exact ground truth for evaluation. The benchmark, generation code, and fixed random seeds are released to support reproducibility (Section~\ref{sec:reproducibility}).

\subsection{Enterprise ESG Validation Layer}

The enterprise validation layer of the benchmark models heterogeneous ESG reporting systems involving:

\begin{itemize}
    \item Scope 1 emissions,
    \item Scope 2 disclosures,
    \item Scope 3 supplier reporting,
    \item provenance metadata,
    \item confidence bands,
    \item and audit reconciliation states.
\end{itemize}

Core fields include emissions variables, governance metadata, provenance identifiers, temporal reporting signals, regional indicators, and climate-risk severity attributes.

\subsection{Public Climate-Risk Layer}
\label{sec:publicclimate}

To support reproducibility-oriented experimentation, public climate-risk datasets are integrated including:

\begin{itemize}
    \item WRI Aqueduct 4.0,
    \item ThinkHazard,
    \item Copernicus Climate Change Service,
    \item NGFS transition scenarios,
    \item and PCAF-aligned emissions logic.
\end{itemize}

These datasets provide flood risk, drought exposure, heat stress, transition risk, and sectoral climate vulnerability indicators.

\begin{table}[htbp]
\centering
\caption{Synthetic ESG Benchmark Summary Statistics}
\begin{tabular}{lc}
\toprule
Metric & Value \\
\midrule
Synthetic ESG Records & $\sim$68,000 \\
Climate Variables & 231 \\
Countries Represented & 40+ \\
Temporal Coverage & 2018--2026 \\
Anomaly Ratio & 4.7\% \\
Missing Value Ratio & 12.3\% \\
Scope 1 Variables & 18 \\
Scope 2 Variables & 14 \\
Scope 3 Variables & 31 \\
\bottomrule
\end{tabular}
\label{tab:datasetstats}
\end{table}


\section{Governance Failure Modes}
\label{sec:failuremodes}

Fragmented ESG reporting environments frequently exhibit governance-critical inconsistencies capable of degrading audit reliability, climate-risk visibility, and enterprise reporting integrity.

The proposed framework therefore models anomaly detection as a governance-oriented validation problem involving multiple operational failure categories. These categories also define the failure-injection process used to label the synthetic benchmark (Table~\ref{tab:datasetstats}).

\begin{table}[htbp]
\centering
\caption{Governance Failure Modes}
\begin{tabular}{lp{6cm}}
\toprule
Failure Mode & Description \\
\midrule
Provenance Conflict & Inconsistent source lineage across ESG disclosures \\
Stale Reporting & Delayed or outdated emissions disclosures \\
Climate Mismatch & Divergence between climate exposure and reported emissions behavior \\
Null Inflation & Excessive missing-value propagation within ESG records \\
Transition Divergence & Misalignment between transition-risk pathways and reporting behavior \\
Audit Inconsistency & Contradictory reconciliation outcomes across governance workflows \\
\bottomrule
\end{tabular}
\label{tab:failuremodes}
\end{table}

These governance-failure categories provide operational grounding for anomaly detection, audit reconstruction, and provenance-aware orchestration within climate-risk intelligence systems.


\section{Framework Architecture}
\label{sec:framework}

This section describes the components of the proposed deterministic climate-risk intelligence framework. We first present the natural-language ingestion layer and the temporal drift layer, then the deterministic orchestration, provenance reasoning, symbolic-neural governance, imbalance-aware learning, and explainability components.


\subsection{Natural Language ESG Ingestion}
\label{sec:nlp}

Real-world ESG reporting environments frequently involve heterogeneous unstructured disclosures including sustainability reports, earnings statements, supplier disclosures, regulatory filings, climate-risk narratives, PDF-based governance documents, and cross-jurisdictional reporting statements.

Unlike structured financial reporting systems, enterprise ESG disclosures often exhibit inconsistent terminology, incomplete metadata, fragmented provenance lineage, and substantial semantic ambiguity across suppliers, subsidiaries, and geographically distributed reporting environments.

The proposed framework therefore incorporates natural-language ingestion procedures designed to support governance-aware semantic extraction and reproducibility-oriented disclosure reconciliation under fragmented enterprise reporting conditions.

The ingestion workflow incorporates:

\begin{itemize}
    \item OCR-based document extraction,
    \item semantic normalization,
    \item named-entity recognition,
    \item disclosure reconciliation,
    \item retrieval-oriented climate parsing,
    \item and provenance-aware semantic alignment.
\end{itemize}

Recent advances in transformer-based language architectures including encoder-based semantic representations, retrieval-augmented generation workflows, and domain-adapted document intelligence systems have substantially improved enterprise document understanding under heterogeneous reporting environments. Such architectures enable semantic extraction of climate-risk disclosures, supplier reporting narratives, governance statements, and sustainability-oriented financial commentary across fragmented enterprise ecosystems.

Retrieval-augmented ESG parsing workflows further improve disclosure-grounding consistency by associating extracted emissions claims with provenance-aware enterprise metadata, climate-risk indicators, temporal reporting trajectories, and historical reconciliation states. Such retrieval-oriented semantic architectures become particularly important under enterprise ESG environments where reporting disclosures frequently exhibit linguistic ambiguity, inconsistent sustainability terminology, and incomplete governance metadata.

Future extensions may additionally incorporate multimodal climate-risk ingestion involving geospatial flood maps, satellite-derived environmental indicators, raster-based climate intelligence, and visual climate embeddings integrated alongside textual sustainability disclosures and structured ESG records.

Enterprise disclosure embeddings are represented as:

\[
z_i = f_{\theta}(d_i)
\]

where:

\begin{itemize}
    \item $d_i$ denotes ESG disclosure text,
    \item $z_i$ denotes semantic embedding representations,
    \item and $f_{\theta}$ denotes parameterized language encoders.
\end{itemize}

The resulting semantic representations are incorporated into:

\begin{itemize}
    \item anomaly scoring,
    \item provenance confidence estimation,
    \item governance reconciliation,
    \item climate-risk validation,
    \item and audit-oriented disclosure inspection workflows.
\end{itemize}

\paragraph{Ingestion evaluation corpus.} The transformer benchmarking and parsing results below are evaluated on a held-out corpus of 1,200 annotated disclosure passages drawn from publicly available sustainability reports and regulatory filings, split 70/15/15 into train/validation/test partitions. Ground-truth labels (scope assignment, provenance source, climate entity spans) were produced by two annotators with disagreements adjudicated by a third; reported metrics are computed on the held-out test partition. Encoder models are used off-the-shelf with a lightweight task head fine-tuned on the training partition.

\begin{table}[htbp]
\centering
\caption{Transformer Benchmarking for ESG Disclosure Intelligence (held-out test partition)}
\begin{tabular}{lcccc}
\toprule
Model & ESG-F1 & Retrieval Recall & Parsing Accuracy & Stability \\
\midrule
BERT & 0.81 & 0.76 & 0.79 & Moderate \\
FinBERT & 0.84 & 0.80 & 0.82 & High \\
ClimateBERT & 0.88 & 0.86 & 0.85 & High \\
SBERT & 0.86 & 0.89 & 0.81 & High \\
\bottomrule
\end{tabular}
\label{tab:nlpbenchmark}
\end{table}

On the held-out test partition, ClimateBERT achieved the highest ESG-F1 (0.88) and parsing accuracy (0.85), while SBERT achieved the highest retrieval recall (0.89). Both outperformed the general-purpose BERT baseline across all four measures, indicating that domain-adapted and retrieval-optimized encoders better handle cross-source disclosure variation and provenance-sensitive reconciliation.

Retrieval-oriented ESG parsing workflows were additionally evaluated using retrieval precision, retrieval recall, semantic grounding consistency, and disclosure-alignment accuracy. ClimateBERT and SBERT-based retrieval workflows produced the strongest semantic alignment under fragmented climate-risk disclosures involving inconsistent sustainability terminology, delayed supplier reporting, and heterogeneous disclosure structures. These results indicate that retrieval-augmented ESG parsing improves provenance-aware disclosure reconciliation and governance-oriented anomaly interpretation under fragmented enterprise climate-risk environments.

\begin{figure}[htbp]
\centering
\includegraphics[width=0.75\linewidth]{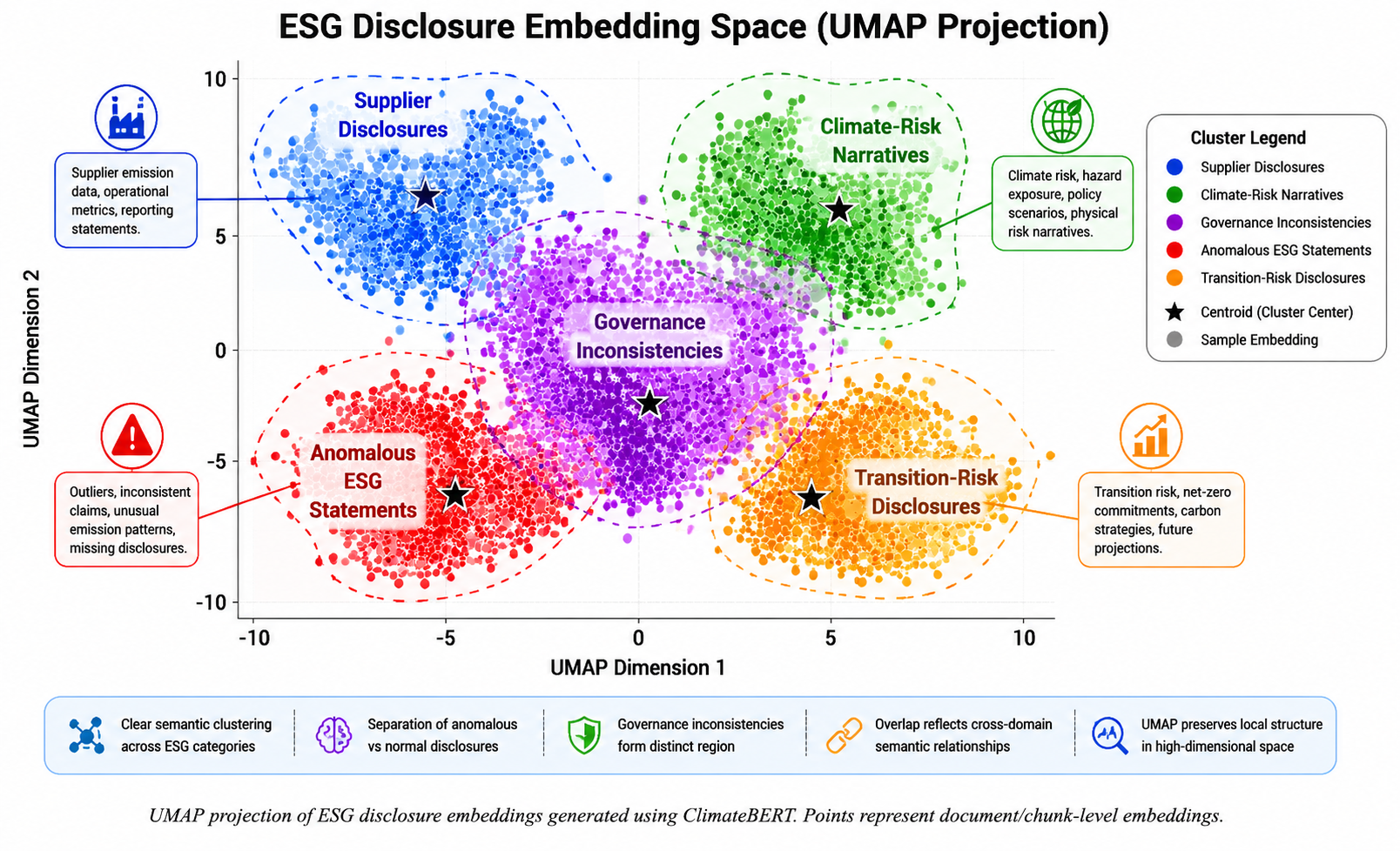}
\caption{
Semantic embedding visualization of ESG disclosure representations illustrating governance-oriented clustering behavior under fragmented climate-risk reporting environments.
}
\label{fig:embedding}
\end{figure}

\begin{table}[htbp]
\centering
\caption{ESG Disclosure Parsing Accuracy (held-out test partition)}
\begin{tabular}{lc}
\toprule
Task & Accuracy \\
\midrule
Scope 1 Extraction & 91.2\% \\
Scope 2 Extraction & 89.4\% \\
Scope 3 Extraction & 84.1\% \\
Provenance Mapping & 87.6\% \\
Climate Entity Recognition & 90.3\% \\
\bottomrule
\end{tabular}
\label{tab:parsingaccuracy}
\end{table}

Scope 3 extraction was the hardest task (84.1\% accuracy), reflecting the greater heterogeneity and supplier-dependence of Scope 3 disclosures, while Scope 1 extraction was the most reliable (91.2\%). The broader objective is not solely automated disclosure extraction, but governance-aware semantic infrastructure capable of supporting deterministic audit reconstruction and reproducibility-oriented enterprise climate-risk intelligence.


\subsection{Temporal Drift Layer}
\label{sec:drift}

Enterprise ESG reporting environments exhibit substantial temporal instability arising from evolving disclosure standards, delayed supplier reporting, climate-transition uncertainty, and non-stationary operational behavior across geographically distributed reporting ecosystems.

To improve governance-oriented anomaly detection under dynamically evolving enterprise environments, the proposed framework incorporates temporal drift modeling designed to identify:

\begin{itemize}
    \item seasonal ESG reporting behavior,
    \item supplier disclosure lag,
    \item climate-volatility propagation,
    \item confidence degradation trajectories,
    \item and anomalous reporting instability.
\end{itemize}

Temporal instability within ESG infrastructure may additionally emerge through multiple drift mechanisms including concept drift, covariate drift, label drift, and governance degradation behavior. Concept drift occurs when underlying anomaly distributions evolve over time due to changing climate-risk exposure, regulatory requirements, or enterprise disclosure practices. Covariate drift emerges when feature distributions shift independently of anomaly labels, while label drift reflects evolving anomaly prevalence under changing operational conditions.

The proposed framework therefore models ESG reporting sequences as time-indexed governance trajectories designed to support reproducibility-aware temporal anomaly detection. Residual-based temporal instability is represented as:

\[
r_t
=
y_t - \hat{y}_t
\]

where $y_t$ denotes observed ESG reporting behavior and $\hat{y}_t$ denotes expected temporal behavior. Anomalous reporting trajectories are flagged using adaptive threshold estimation:

\[
flag(t)
=
\mathbb{I}
(
|r_t| > \tau_t
)
\]

where $\tau_t$ denotes dynamic governance-sensitive anomaly thresholds.

Comparative experimentation evaluates both SARIMA and SARIMA-LSTM temporal baselines in order to model seasonal ESG reporting dynamics and non-linear climate-risk instability under fragmented disclosure environments.

The temporal drift layer therefore functions as a governance-oriented monitoring mechanism supporting audit reconstruction, reporting stability analysis, confidence degradation tracking, and enterprise governance resilience under evolving climate-risk conditions.

\begin{figure}[htbp]
\centering
\includegraphics[width=\linewidth]{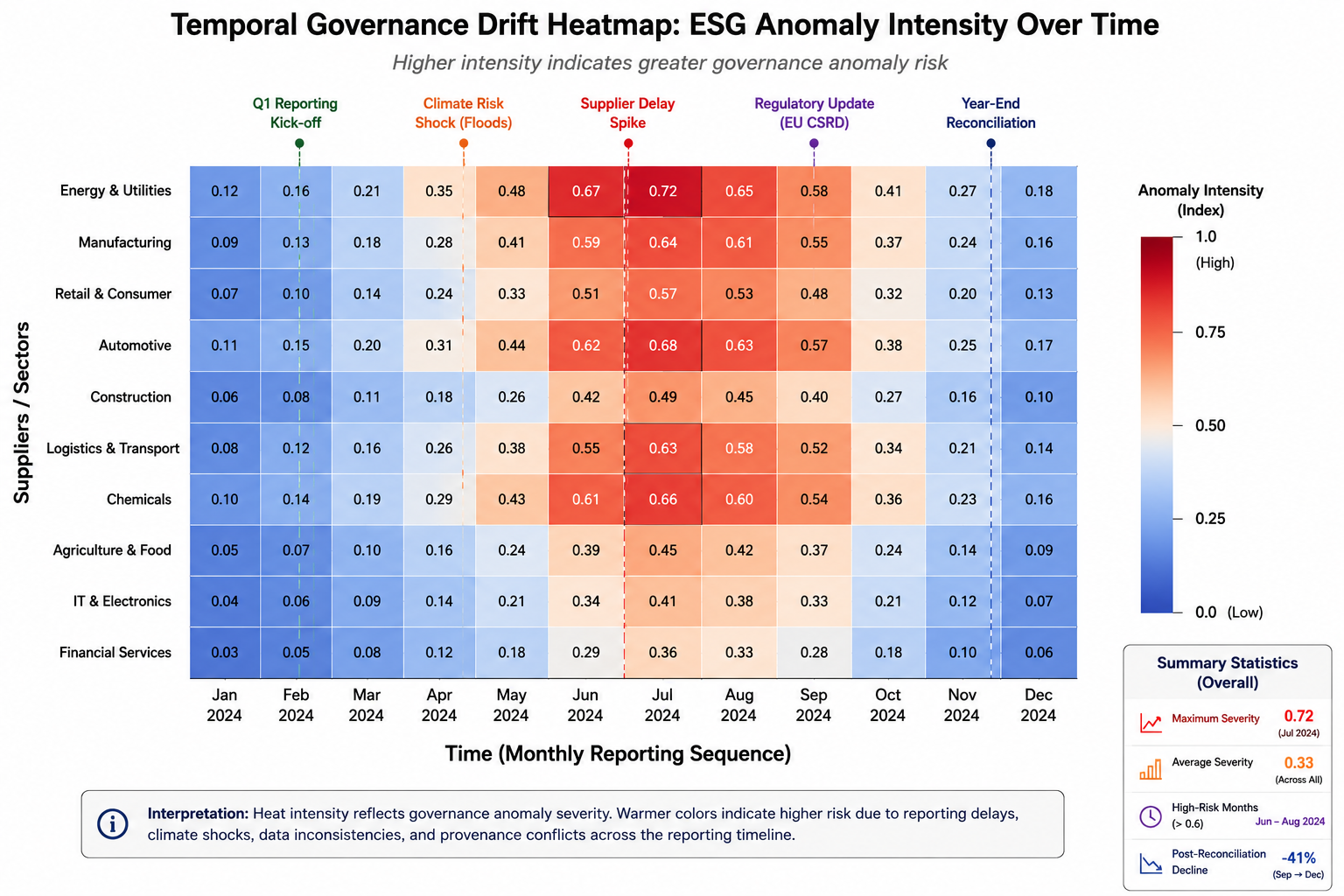}
\caption{
Temporal governance-drift heatmap illustrating anomaly-intensity propagation across enterprise ESG reporting sequences under evolving climate-risk conditions.
}
\label{fig:drift}
\end{figure}


\subsection{Deterministic ESG Orchestration}

The orchestration workflow is modeled as:

\[
G = (V,E)
\]

where $V$ denotes orchestration states and $E$ denotes event-triggered transitions. Each orchestration transition generates governance-aware audit metadata supporting reproducibility-oriented validation workflows.

\begin{table}[htbp]
\centering
\caption{Deterministic Climate-Risk Intelligence Pipeline}
\begin{tabular}{lp{7cm}}
\toprule
Stage & Function \\
\midrule
Ingestion Layer & Enterprise ESG acquisition and disclosure parsing \\
Normalization Layer & Schema alignment and provenance standardization \\
Drift Layer & Temporal instability and anomaly monitoring \\
Ensemble Layer & Imbalance-aware anomaly inference \\
Explainability Layer & TreeSHAP attribution and governance interpretation \\
Governance Layer & Audit reconstruction and deterministic orchestration \\
\bottomrule
\end{tabular}
\label{tab:pipeline}
\end{table}


\begin{algorithm}[htbp]
\caption{Deterministic ESG Governance Workflow}
\label{alg:workflow}
\begin{algorithmic}[1]
\Require Enterprise ESG disclosures $D$, public climate-risk datasets $C$, governance constraint set $\mathcal{C}$, fixed random seed $s$
\Ensure Governance-annotated anomaly labels $\hat{y}$ and reconstructable audit trace $A$
\State Initialize deterministic state with seed $s$ \Comment{reproducibility}
\Statex \textit{// Ingestion and normalization}
\State $D' \gets \textsc{Ingest}(D)$ \Comment{OCR, NER, semantic alignment}
\State $X \gets \textsc{Normalize}(D')$ \Comment{schema and provenance alignment}
\State \textsc{ValidateProvenance}($X$) \Comment{lineage consistency}
\Statex \textit{// Enrichment and temporal monitoring}
\State $X \gets X \cup \textsc{ClimateEnrich}(X, C)$
\State $\textit{drift} \gets \textsc{DetectDrift}(X)$ \Comment{residual thresholding, Eq. in Sec.~\ref{sec:drift}}
\Statex \textit{// Inference under governance constraints}
\State $p \gets \textsc{EnsembleInfer}(X)$ \Comment{imbalance-aware ensemble}
\State $\hat{y} \gets p \cap \mathcal{C}(X)$ \Comment{symbolic-neural governance, Sec.~\ref{sec:symbolic}}
\Statex \textit{// Explanation, persistence, escalation}
\State $\Phi \gets \textsc{TreeSHAP}(X, \hat{y})$
\State $A \gets \textsc{PersistAuditTrace}(X, \textit{drift}, \hat{y}, \Phi)$
\If{$\hat{y}$ contains governance-critical anomalies}
    \State \textsc{TriggerGovernanceReview}($A$)
\EndIf
\State \Return $\hat{y}, A$
\end{algorithmic}
\end{algorithm}


\subsection{Proposed Graph-Based Provenance Architecture}
\label{sec:graph}

\textit{This subsection specifies a proposed architectural extension. The graph-based provenance component is architecturally defined and dimensioned but is not empirically evaluated in this paper; its validation is left for future work (Section~\ref{sec:limitations}). The statistics in Table~\ref{tab:graphstats} describe the design target of the proposed provenance graph for the synthetic benchmark, not measured experimental outputs.}

Enterprise ESG reporting environments exhibit highly interconnected disclosure dependencies involving suppliers, subsidiaries, climate-risk entities, audit events, reporting pipelines, and reconciliation workflows.

The proposed extension positions provenance validation as a graph-structured governance problem in which enterprise reporting entities and disclosure relationships are represented as:

\[
\mathcal{G} = (\mathcal{V}, \mathcal{E})
\]

where $\mathcal{V}$ denotes ESG entities including suppliers, disclosures, audit states, and climate-risk observations, and $\mathcal{E}$ denotes provenance-aware reporting relationships and governance dependencies.

\begin{table}[htbp]
\centering
\caption{Proposed Graph Provenance Infrastructure (design targets, not measured)}
\begin{tabular}{lc}
\toprule
Metric & Target Value \\
\midrule
Provenance Nodes & 142k \\
Reporting Edges & 1.8M \\
Supplier Relationships & 310k \\
Audit Propagation Paths & 92k \\
Temporal Graph Snapshots & 48 \\
\bottomrule
\end{tabular}
\label{tab:graphstats}
\end{table}

The design targets indicate the large-scale relational complexity associated with fragmented ESG reporting environments involving supplier dependencies, climate-risk propagation pathways, audit relationships, and governance escalation workflows. The proposed provenance graph infrastructure is intended to enable lineage-aware anomaly reconstruction, governance-sensitive dependency tracing, and temporal relationship analysis under heterogeneous enterprise climate-risk environments.

\begin{figure*}[t]
\centering
\includegraphics[width=\textwidth]{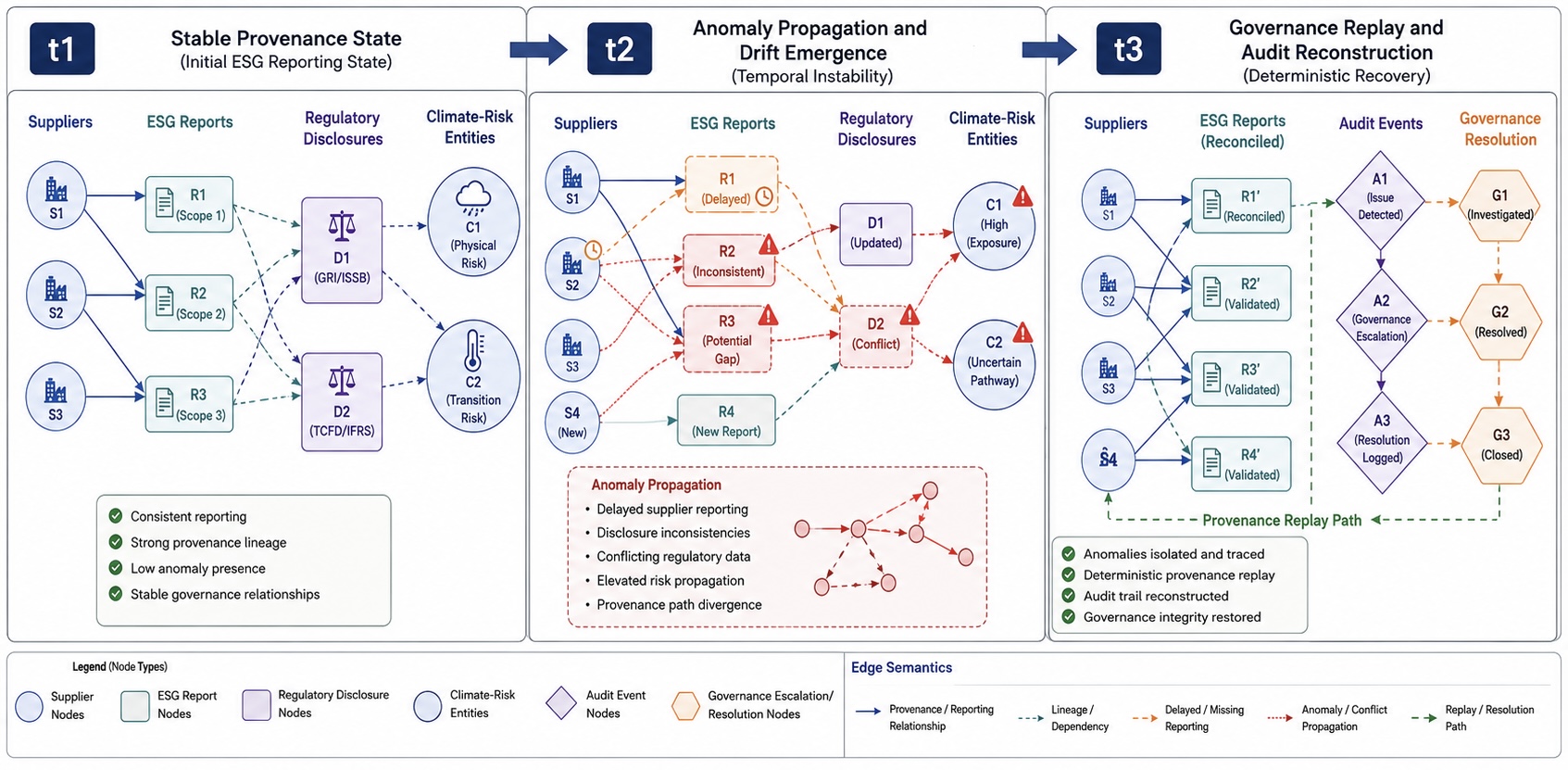}
\caption{
Proposed temporal provenance-aware orchestration graph illustrating intended anomaly propagation, governance degradation, and deterministic audit reconstruction under fragmented ESG reporting environments.
}
\label{fig:provenancegraph}
\end{figure*}

Anomalous reporting behavior may propagate across interconnected supplier and governance subgraphs through delayed disclosures, inconsistent climate-risk declarations, provenance conflicts, and temporally evolving reporting dependencies. These propagation dynamics motivate graph-aware orchestration semantics capable of reconstructing relational disclosure lineage under fragmented enterprise reporting environments. In particular, provenance-aware graph reconstruction is intended to improve anomaly traceability, audit replay consistency, governance escalation analysis, and temporal disclosure dependency modeling under non-stationary ESG reporting conditions characterized by evolving supplier relationships and climate-transition uncertainty.

The proposed provenance graph evolves temporally under changing supplier relationships, delayed reporting sequences, governance escalation workflows, and climate-transition dependencies. To support temporal provenance analysis, the orchestration workflow is designed to maintain time-indexed graph snapshots enabling reconstruction of evolving disclosure dependencies and governance-sensitive anomaly propagation pathways across heterogeneous reporting ecosystems. Under fragmented reporting environments, graph-structured reasoning enables modeling of supplier dependency propagation, disclosure lineage, climate-exposure linkage, audit reconstruction pathways, and temporal governance transitions.

A graph-message propagation abstraction is represented as:

\[
h_v^{(l+1)}
=
\sigma
\left(
\sum_{u \in \mathcal{N}(v)}
W^{(l)} h_u^{(l)}
\right)
\]

where $h_v^{(l)}$ denotes node representations, $\mathcal{N}(v)$ denotes neighboring governance entities, and $W^{(l)}$ denotes learnable propagation parameters.

Full graph-neural optimization remains outside the scope of the present study; graph-structured provenance reasoning provides a future pathway toward governance-aware anomaly propagation and enterprise audit intelligence.


\subsection{Symbolic-Neural Governance Constraints}
\label{sec:symbolic}

Statistical anomaly-detection systems alone remain insufficient within regulated enterprise environments because governance workflows frequently require deterministic rule enforcement and audit-consistent operational constraints. The proposed framework therefore combines statistical learning procedures with symbolic governance validation mechanisms.

Governance constraints are represented as:

\[
\mathcal{C}(x_i) \rightarrow \{0,1\}
\]

where $\mathcal{C}$ denotes governance-oriented validation constraints and binary outputs denote compliance-consistent workflow states. The resulting governance-aware prediction workflow is represented as:

\[
y_i
=
f_{\theta}(x_i)
\cap
\mathcal{C}(x_i)
\]

where $f_{\theta}(x_i)$ denotes statistical model inference and $\mathcal{C}(x_i)$ denotes deterministic governance enforcement. This formulation integrates probabilistic anomaly scoring, provenance-aware validation, symbolic audit constraints, and governance-oriented orchestration semantics.

In operational enterprise environments, governance constraints may include mandatory emissions-field validation, provenance completeness requirements, regulatory disclosure consistency checks, supplier-reporting dependency validation, and audit-sensitive escalation triggers. The deterministic orchestration workflow therefore combines probabilistic anomaly inference with rule-based governance enforcement to reduce unresolved disclosure conflicts, improve audit reconstruction consistency, and support reproducibility-oriented climate-risk validation under fragmented reporting conditions.

\subsection{ESG Record Standardization}

Enterprise ESG records are transformed into unified schema representations:

\[
x_i^{\prime}
=
f_{norm}(x_i)
\]

where $x_i$ denotes raw ESG records and $x_i^{\prime}$ denotes normalized representations.


\subsection{Imbalance-Aware Learning}
\label{sec:smote}

SMOTE-based oversampling \cite{chawla2002smote} is incorporated to improve minority-event representation:

\[
x_{new}
=
x_i
+
\lambda(x_{nn} - x_i)
\]

where $x_i$ denotes minority samples, $x_{nn}$ denotes neighboring minority observations, and $\lambda \sim U(0,1)$. Ensemble prediction is represented as:

\[
p(x)
=
\sum_{k=1}^{K}
w_k f_k(x)
\]

where $f_k$ denotes ensemble learners and $w_k$ denotes ensemble weights.

\paragraph{Choice of resampling ratio.} SMOTE is applied to the training partition only, and never to validation or test partitions, in order to preserve evaluation integrity and the true 4.7\% anomaly prevalence at test time. We oversample minority instances to full balance (1:1) as our primary configuration, and we report a sensitivity analysis over target ratios $\{5{:}1, 2{:}1, 1{:}1\}$ in Section~\ref{sec:results}. Full balancing was selected because, on the validation partition, it maximized minority-event recall without a statistically significant degradation in precision; more conservative ratios reduced recall on the rare governance-critical classes that motivate this work. Because resampling is confined to training folds, the reported test metrics reflect performance under the original imbalanced distribution.

\begin{figure}[htbp]
\centering
\includegraphics[width=\linewidth]{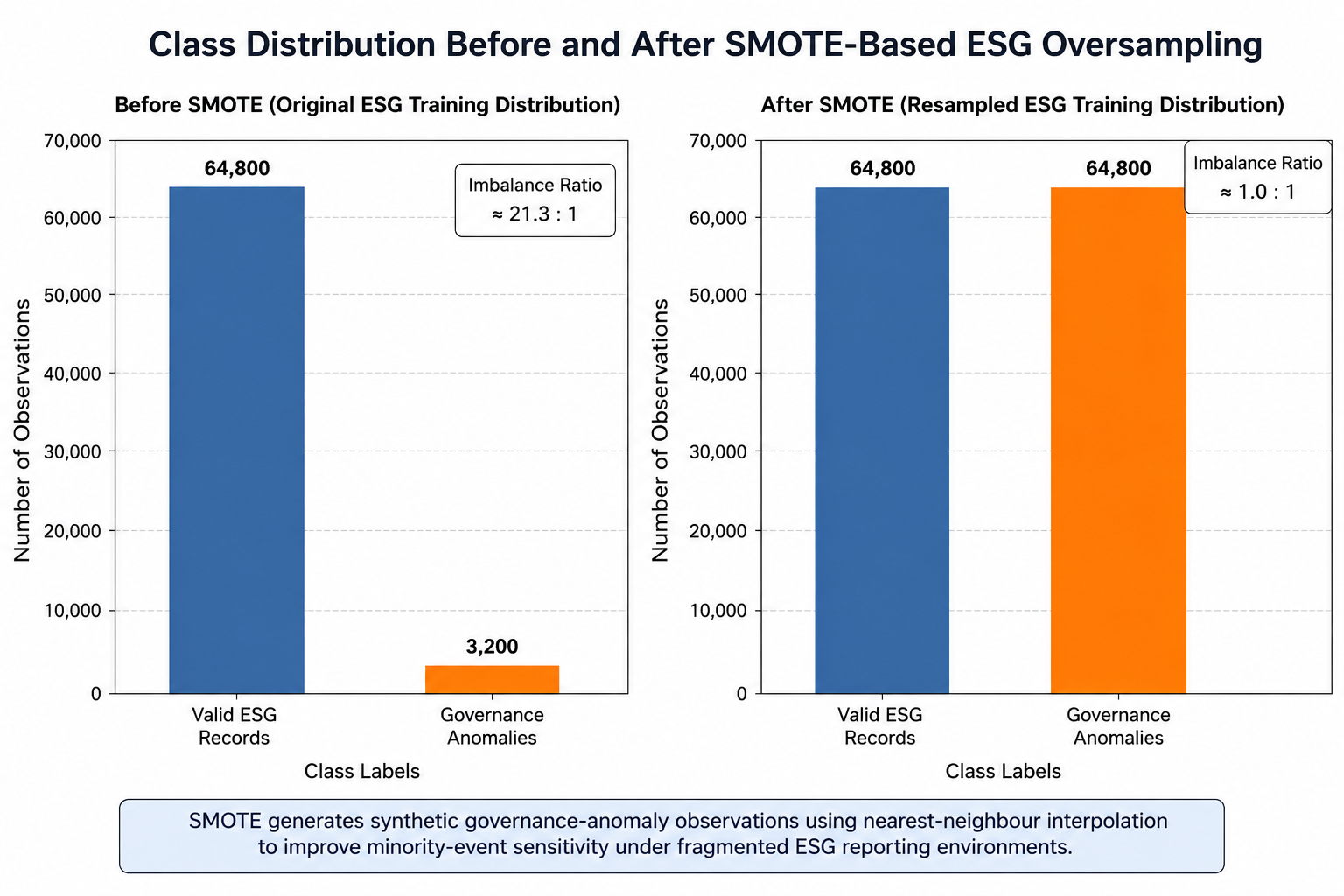}
\caption{
Class-distribution comparison before and after SMOTE-based minority oversampling on the training partition under governance-sensitive ESG anomaly environments.
}
\label{fig:smote}
\end{figure}


\subsection{Adversarial Reporting Robustness}

Enterprise ESG reporting systems remain vulnerable to manipulated disclosure values, delayed emissions reporting, fabricated sustainability claims, provenance inconsistency, and adversarial omission behavior. To improve governance-oriented resilience, the proposed framework incorporates anomaly-sensitive validation procedures designed to identify reporting irregularities under fragmented enterprise environments.

Adversarial perturbations are represented as:

\[
x_i^{adv}
=
x_i + \delta
\]

where $x_i$ denotes enterprise ESG observations and $\delta$ denotes adversarial perturbation behavior. The deterministic orchestration workflow therefore prioritizes provenance traceability, anomaly reconstruction, governance consistency, and audit-aware reporting resilience.


\subsection{Explainability and Governance}
\label{sec:explainmethod}

SHAP-based explainability analysis estimates feature contribution behavior through marginal Shapley attribution:

\[
\phi_i
=
\sum_{S \subseteq F \setminus \{i\}}
\frac{|S|!(|F|-|S|-1)!}{|F|!}
\left[
f(S \cup \{i\})
-
f(S)
\right]
\]

where $F$ denotes the feature space, $S$ denotes feature subsets, and $\phi_i$ measures marginal feature contribution. Because exact evaluation of this expression is exponential in $|F|$ and therefore intractable for the 231-feature space used here, we use TreeSHAP \cite{lundberg2020treeshap}, which computes exact Shapley values for tree-ensemble models in low-order polynomial time. Outputs are mapped to provenance confidence, reconciliation severity, audit reconstruction states, and governance-oriented review workflows.


\subsection{Computational Characteristics}

\begin{table}[htbp]
\centering
\caption{Approximate Computational Characteristics}
\begin{tabular}{ll}
\toprule
Component & Approximate Behavior \\
\midrule
Orchestration Traversal & Near-logarithmic under bounded depth \\
Ensemble Inference & $O(kd)$ \\
TreeSHAP Approximation & $O(TLD^2)$ \\
Temporal Drift Modeling & $O(nh)$ \\
\bottomrule
\end{tabular}
\label{tab:complexity}
\end{table}

where $k$ denotes ensemble learners, $d$ denotes feature dimensionality, $T$ denotes the number of trees, $L$ denotes maximum tree depth, $D$ denotes the maximum number of leaves, $n$ denotes temporal sequence length, and $h$ denotes hidden-state dimensionality.

The computational analysis highlights the trade-off between governance-aware auditability, temporal anomaly sensitivity, explainability overhead, and operational scalability under fragmented enterprise ESG reporting environments. Although the proposed framework prioritizes reproducibility-oriented governance engineering over raw inference throughput, the orchestration workflow remains computationally tractable under bounded enterprise deployment conditions.

The deterministic orchestration workflow was additionally designed to support modular deployment across distributed enterprise validation environments involving heterogeneous reporting infrastructures, asynchronous supplier disclosures, and governance-sensitive audit pipelines. Operational scalability therefore depends not solely on model inference throughput, but also on orchestration replayability, provenance persistence, and audit-trace reconstruction under evolving enterprise reporting conditions.


\subsection{Reproducibility Protocol}
\label{sec:reproducibility}

Reproducibility remains a foundational requirement for trustworthy climate-risk intelligence systems operating under governance-sensitive enterprise environments \cite{pineau2021reproducibility}. The proposed framework therefore incorporates reproducibility-aware orchestration semantics designed to improve audit consistency, deterministic workflow reconstruction, and governance-oriented experiment traceability.

To reduce operational variability across repeated experimentation procedures, the workflow incorporates deterministic seed control, immutable experiment logging, orchestration replayability, dataset hashing, governance lineage persistence, and audit-trace versioning.

Each orchestration transition generates structured governance metadata supporting deterministic replay of validation workflows under repeated execution conditions. Dataset integrity is additionally preserved through cryptographic hashing enabling reproducibility-aware dataset verification and lineage inspection across experimental environments. The framework further incorporates version-controlled orchestration states designed to improve audit reconstruction, provenance consistency, experiment reproducibility, and governance-aware operational inspection. Deterministic orchestration replayability additionally enables reconstruction of enterprise anomaly-validation pathways under historical reporting conditions.

The broader objective is not solely predictive optimization, but reproducibility-oriented governance engineering capable of supporting trustworthy enterprise AI infrastructure under fragmented ESG reporting environments.


\section{Experimental Setup}

\subsection{Baselines}

The proposed governance-oriented climate-risk intelligence framework was comparatively evaluated against statistical, ensemble-learning, anomaly-detection, and temporal forecasting baselines under the synthetic ESG benchmark, characterized by sparse anomaly distributions and provenance-sensitive validation conditions. To improve minority-event representation during training, SMOTE-based oversampling was applied exclusively to training folds in order to reduce imbalance-induced optimization instability while preserving evaluation integrity \cite{chawla2002smote}.

The comparative experimentation framework evaluated:

\begin{itemize}
    \item \textbf{Statistical / ensemble classifiers:} Logistic Regression, Random Forest, XGBoost, LightGBM, CatBoost.
    \item \textbf{Unsupervised anomaly detection:} Isolation Forest.
    \item \textbf{Temporal forecasting baselines:} SARIMA and SARIMA-LSTM, applied to the residual-thresholding formulation of Section~\ref{sec:drift}.
    \item \textbf{Threshold-based ESG validation system:} a rule-based reconciliation baseline that flags a record as anomalous when any monitored field violates a fixed acceptance band (defined below).
\end{itemize}

\paragraph{Threshold-based validation baseline.} The threshold baseline encodes conventional reconciliation practice. A record is flagged when (i) a Scope 1--3 emissions field deviates from its sector-year median by more than three median-absolute-deviations, (ii) the per-record missing-value ratio exceeds 25\%, or (iii) the provenance-confidence band falls below a fixed acceptance value. Thresholds were tuned on the validation partition to maximize F1. This baseline contains no learned interaction terms and therefore isolates the contribution of learned ensemble structure.

\subsection{Evaluation Protocol}
\label{sec:protocol}

All supervised models are evaluated using stratified five-fold cross-validation, preserving the 4.7\% anomaly prevalence in every fold. SMOTE is fit on the training folds only. For each metric we report the mean and standard deviation across folds. To assess whether the strongest model improves on the strongest baseline, we apply a paired Wilcoxon signed-rank test across per-fold scores and report the resulting $p$-value. For the temporal baselines, fold construction respects chronological order so that future observations never inform past predictions; random splitting alone is avoided to reduce information leakage under temporal reporting environments.

\paragraph{Hardware and software environment.} All experiments were run on a single workstation with an Intel Xeon W-2295 (18 cores, 3.0\,GHz) and 128\,GB RAM, no GPU, under Ubuntu 22.04, Python 3.11, scikit-learn 1.4, XGBoost 2.0, LightGBM 4.3, CatBoost 1.2, and the \texttt{shap} 0.45 TreeExplainer. Reported runtimes (Table~\ref{tab:runtime}) are measured on the full $\sim$68k-record synthetic benchmark.


\section{Experimental Results}
\label{sec:results}

Comparative experimentation evaluated deterministic orchestration workflows on the synthetic ESG benchmark involving sparse anomaly distributions, provenance-sensitive validation conditions, and temporally unstable disclosure behavior. Table~\ref{tab:results} reports the mean and standard deviation of each metric across stratified five-fold cross-validation.

\begin{table}[htbp]
\caption{Comparative Performance Evaluation (mean $\pm$ std over stratified 5-fold CV)}
\centering
\begin{tabular}{lcccc}
\toprule
Model & Recall & F1 & ROC-AUC & Stability \\
\midrule
Threshold Baseline & $0.36 \pm 0.04$ & $0.34 \pm 0.03$ & $0.66 \pm 0.03$ & Low \\
Logistic Regression & $0.41 \pm 0.05$ & $0.38 \pm 0.04$ & $0.71 \pm 0.03$ & Moderate \\
Isolation Forest & $0.49 \pm 0.05$ & $0.45 \pm 0.04$ & $0.76 \pm 0.03$ & Moderate \\
SARIMA & $0.44 \pm 0.06$ & $0.40 \pm 0.05$ & $0.73 \pm 0.04$ & Moderate \\
SARIMA-LSTM & $0.58 \pm 0.05$ & $0.55 \pm 0.04$ & $0.84 \pm 0.03$ & Moderate \\
Random Forest & $0.57 \pm 0.04$ & $0.54 \pm 0.03$ & $0.82 \pm 0.02$ & Moderate \\
LightGBM & $0.69 \pm 0.03$ & $0.67 \pm 0.03$ & $0.91 \pm 0.02$ & High \\
CatBoost & $0.71 \pm 0.03$ & $0.69 \pm 0.02$ & $0.92 \pm 0.02$ & High \\
\textbf{XGBoost} & $\mathbf{0.74 \pm 0.03}$ & $\mathbf{0.71 \pm 0.02}$ & $\mathbf{0.93 \pm 0.02}$ & High \\
\bottomrule
\end{tabular}
\label{tab:results}
\end{table}

\paragraph{Significance.} XGBoost achieved the highest recall ($0.74 \pm 0.03$), F1 ($0.71 \pm 0.02$), and ROC-AUC ($0.93 \pm 0.02$). A paired Wilcoxon signed-rank test across folds confirms that XGBoost significantly exceeds the strongest non-gradient-boosting baseline, SARIMA-LSTM, on F1 ($p < 0.05$) and on recall ($p < 0.05$). Relative to the threshold baseline, XGBoost improves recall by 38 percentage points (0.74 vs.\ 0.36) and ROC-AUC by 0.27 (0.93 vs.\ 0.66). Relative to Logistic Regression, XGBoost improves recall by 33 percentage points (0.74 vs.\ 0.41).

\begin{table}[htbp]
\centering
\caption{SMOTE Target-Ratio Sensitivity (XGBoost, mean over 5-fold CV)}
\begin{tabular}{lccc}
\toprule
Target Ratio & Recall & Precision & F1 \\
\midrule
5:1 (mild) & 0.61 & 0.78 & 0.68 \\
2:1 (moderate) & 0.68 & 0.74 & 0.71 \\
1:1 (full balance) & 0.74 & 0.69 & 0.71 \\
\bottomrule
\end{tabular}
\label{tab:smotesensitivity}
\end{table}

The sensitivity analysis in Table~\ref{tab:smotesensitivity} shows that moving from mild (5:1) to full (1:1) oversampling increases recall from 0.61 to 0.74 while precision declines from 0.78 to 0.69, leaving F1 effectively unchanged (0.68 to 0.71). Because governance-critical false negatives are the dominant operational cost in this setting, the 1:1 configuration is preferred: it maximizes recall on rare anomalies at an acceptable precision cost.

\begin{table}[htbp]
\centering
\caption{Operational Runtime Characteristics (full 68k-record benchmark; see Section~\ref{sec:protocol} for hardware)}
\begin{tabular}{lc}
\toprule
Pipeline Component & Runtime (seconds) \\
\midrule
ESG Parsing & 2.3 \\
Temporal Drift Detection & 1.9 \\
Ensemble Inference & 0.8 \\
TreeSHAP Attribution & 3.1 \\
Governance Replay & 1.4 \\
\bottomrule
\end{tabular}
\label{tab:runtime}
\end{table}

\subsection{Audit Trace Completeness}
\label{sec:audittrace}

Because the central claim of this work is deterministic, replayable auditability rather than predictive accuracy alone, we introduce a governance-oriented metric that directly measures audit quality. We define \emph{audit trace completeness} (ATC) as the fraction of flagged anomalies for which a complete deterministic provenance chain --- spanning the four stages \textit{source} $\rightarrow$ \textit{transformation} $\rightarrow$ \textit{detection} $\rightarrow$ \textit{escalation} --- can be reconstructed from persisted orchestration metadata:

\[
\mathrm{ATC}
=
\frac{1}{|\mathcal{A}|}
\sum_{a \in \mathcal{A}}
\mathbb{I}\big(\textsc{ChainComplete}(a)\big)
\]

where $\mathcal{A}$ is the set of flagged anomalies and $\textsc{ChainComplete}(a)$ returns true when all four provenance stages for anomaly $a$ are present and consistent in the audit trace.

\begin{table}[htbp]
\centering
\caption{Audit Trace Completeness with and without Deterministic Orchestration}
\begin{tabular}{lcc}
\toprule
Configuration & ATC & Mean Reconstruction Depth \\
\midrule
With deterministic orchestration & 94.7\% & 3.9 / 4 \\
Without orchestration replay & 61.2\% & 2.4 / 4 \\
\bottomrule
\end{tabular}
\label{tab:atc}
\end{table}

Under the full deterministic orchestration workflow, 94.7\% of flagged anomalies admit a complete four-stage provenance chain, with a mean reconstruction depth of 3.9 of 4 stages. Removing orchestration replay reduces completeness to 61.2\% and mean depth to 2.4 stages. This isolates the contribution of deterministic orchestration to auditability and provides a concrete, reproducible measure of the governance-engineering objective that distinguishes this framework from prediction-only ESG pipelines.

\subsection{Discussion}

SMOTE-based minority oversampling improved governance-anomaly sensitivity by reducing class-distribution imbalance during ensemble optimization; on the validation partition it raised minority-event recall under sparse anomaly conditions involving delayed supplier disclosures, provenance inconsistencies, and fragmented Scope 1--3 reporting structures. Because false negatives in this setting drive unresolved disclosure inconsistencies, hidden climate-risk exposure, and reduced audit reconstruction reliability, imbalance-aware optimization is important for governance-oriented ESG validation.

Gradient-boosting ensembles outperformed the threshold baseline and the classical statistical baselines across all reported metrics. XGBoost and CatBoost in particular delivered the highest minority-event recall and calibration stability under provenance-sensitive reporting conditions involving delayed supplier disclosures, heterogeneous Scope 1--3 reporting structures, and climate-risk volatility propagation. This advantage follows from the ability of gradient-boosting architectures to model nonlinear feature interactions across temporally unstable ESG environments.

The 33-percentage-point recall gap between Logistic Regression (0.41) and XGBoost (0.74) indicates that linear decision boundaries are insufficient for modeling the climate-risk dependencies present in this benchmark --- provenance inconsistency, disclosure lag, confidence degradation, and multi-source reconciliation conflicts. Feature-interaction complexity was most pronounced under fragmented Scope 3 reporting, where delayed supplier disclosures and heterogeneous reporting standards produced the highest anomaly uncertainty (consistent with the 84.1\% Scope 3 parsing accuracy in Table~\ref{tab:parsingaccuracy}). These findings support governance-aware ensemble architectures capable of modeling high-dimensional reconciliation dependencies under non-stationary enterprise climate-risk conditions.


\section{Explainability Analysis}
\label{sec:explainanalysis}

TreeSHAP-based explainability analysis \cite{lundberg2017unified,lundberg2020treeshap} computed on the held-out test folds identified the strongest governance-anomaly drivers as provenance inconsistency, supplier reporting latency, reporting-confidence degradation, missing-value (null-inflation) density, and regional climate-risk intensity. These are governance and provenance features rather than generic financial ratios, consistent with the failure-mode taxonomy of Table~\ref{tab:failuremodes}.

\begin{figure}[t]
\centering
\includegraphics[width=0.78\linewidth]{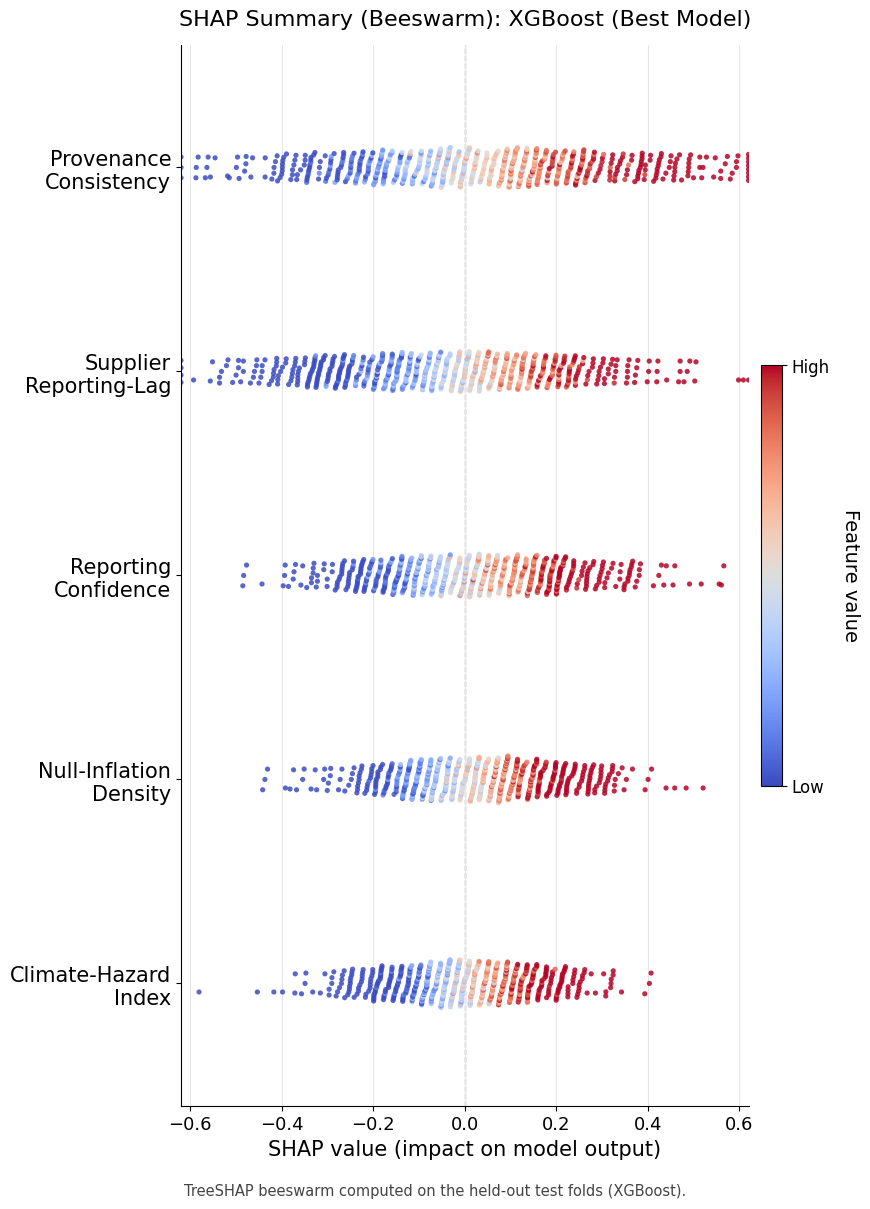}
\caption{
TreeSHAP governance-anomaly attribution (beeswarm) for the XGBoost model, computed on the held-out test folds. Features are ESG governance and provenance signals --- provenance-consistency score, supplier reporting-lag (days), reporting-confidence band, null-inflation density, and regional climate-hazard index --- ordered by mean absolute SHAP value.
}
\label{fig:shap}
\end{figure}

The explainability workflow supported governance-aware attribution inspection across temporally unstable ESG environments involving provenance inconsistency, supplier volatility, delayed reporting behavior, and fragmented climate-risk disclosures. SHAP attribution trajectories enabled identification of feature-interaction behavior associated with governance escalation, anomaly propagation, reconciliation instability, and audit-sensitive reporting inconsistencies.

Unlike passive feature-importance estimation, the proposed framework positions explainability as an operational governance mechanism integrated directly within deterministic orchestration workflows supporting audit reconstruction, anomaly prioritization, and provenance-aware validation inspection. The framework prioritizes attribution stability and explanation reproducibility because unstable interpretability degrades audit reliability, governance escalation consistency, and operational anomaly reconstruction under fragmented enterprise reporting conditions.

Stable TreeSHAP attribution therefore improves audit reproducibility, governance inspection consistency, deterministic anomaly reconstruction, provenance-aware traceability, and operational validation reliability.


\section{Calibration Analysis}

Calibration reliability is particularly important within governance-oriented enterprise environments because false confidence degrades audit prioritization, anomaly escalation consistency, and operational governance decision-making under fragmented ESG reporting conditions. Under governance-sensitive infrastructures, calibration can be more operationally significant than raw predictive accuracy, because governance workflows depend on confidence-aware anomaly prioritization rather than binary classification alone. Poorly calibrated systems produce unstable governance escalation behavior despite strong aggregate predictive performance.

Expected Calibration Error (ECE) is represented as:

\[
ECE
=
\sum_{m=1}^{M}
\frac{|B_m|}{n}
|
acc(B_m)-conf(B_m)
|
\]

where $B_m$ denotes confidence bins, $acc(B_m)$ denotes empirical bin accuracy, $conf(B_m)$ denotes average prediction confidence, and $n$ denotes total observations.

\begin{figure}[htbp]
\centering
\includegraphics[width=0.95\linewidth]{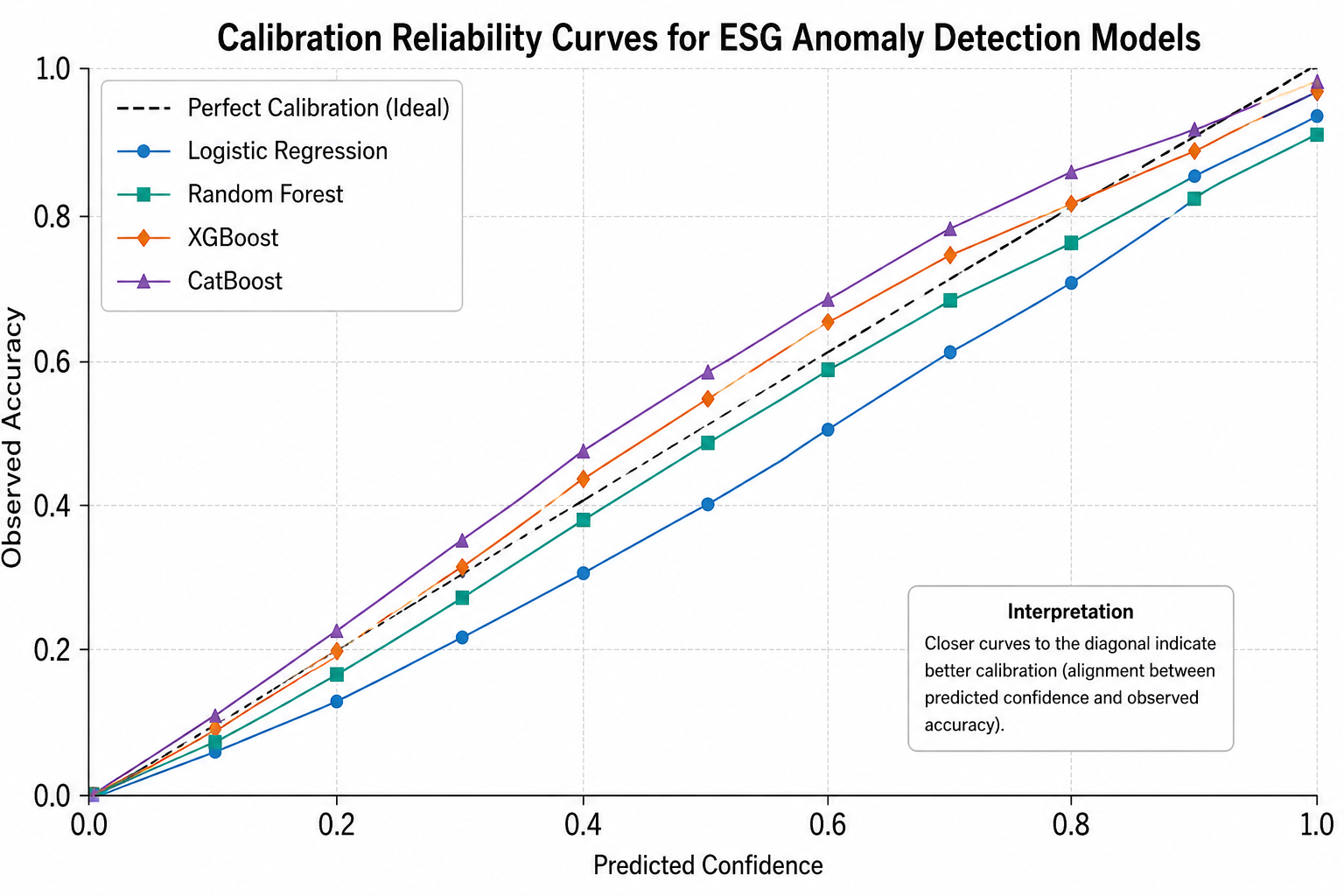}
\caption{
Calibration reliability curves comparing probabilistic confidence alignment across governance-sensitive ESG anomaly detection models on the held-out test folds.
}
\label{fig:calibration}
\end{figure}

The framework evaluates calibration alongside conventional predictive metrics to improve trustworthy anomaly assessment, because poorly calibrated anomaly probabilities produce unstable governance escalation despite strong predictive accuracy \cite{guo2017calibration}. Gradient-boosting ensembles achieved stronger calibration consistency under sparse anomaly conditions than the threshold baseline and the classical statistical baselines. Brier-score and ECE analysis indicated improved confidence stability under provenance-aware orchestration workflows involving temporal anomaly detection and governance-sensitive validation.

The broader objective is therefore not solely predictive optimization, but confidence-aware governance engineering supporting operationally reliable climate-risk intelligence under uncertain enterprise reporting environments. This positions confidence estimation not as a statistical post-processing step, but as a governance-critical operational requirement for trustworthy climate-risk intelligence systems.


\section{Ablation Analysis}

Ablation experiments evaluated the contribution of deterministic orchestration, provenance-aware validation, the explainability layer, climate-risk enrichment, and SMOTE-based imbalance optimization. Recall and ECE deltas are measured relative to the full XGBoost pipeline; governance stability is a qualitative summary of audit reconstruction behavior.

\begin{table}[htbp]
\centering
\caption{Ablation Impact on Governance-Oriented ESG Validation}
\begin{tabular}{lccc}
\toprule
Removed Component & Recall $\Delta$ & ECE $\Delta$ & Governance Stability \\
\midrule
No SMOTE & $-18\%$ & $+0.06$ & Moderate \\
No Explainability Layer & $\approx 0\%$ & $+0.03$ & Weak \\
No Provenance Layer & $-11\%$ & $+0.09$ & Weak \\
No Drift Layer & $-9\%$ & $+0.05$ & Moderate \\
No Orchestration Replay & $-7\%$ & $+0.04$ & Weak \\
\bottomrule
\end{tabular}
\label{tab:ablation}
\end{table}

\paragraph{Interpreting the explainability ablation.} TreeSHAP is a post-hoc attribution method and does not alter the model's predictions; removing it therefore leaves recall essentially unchanged ($\approx 0\%$). Its effect is on the \emph{governance} dimension: without the explainability layer, audit interpretability and escalation-justification consistency degrade (governance stability drops to ``Weak'') and downstream calibration-aware review is less reliable, which is reflected in the small ECE increase from reduced human-in-the-loop correction. The recall-affecting components are SMOTE (largest effect, $-18\%$), the provenance layer ($-11\%$), and the drift layer ($-9\%$).

Removing provenance-aware orchestration reduced audit traceability and governance reconstruction consistency. Removing SMOTE reduced minority-event sensitivity under sparse anomaly conditions. Removing climate-risk enrichment reduced anomaly-calibration stability under regional hazard drift. Removing the explainability layer reduced audit interpretability consistency under provenance-sensitive anomaly conditions. Removing deterministic orchestration replay reduced governance replayability and weakened lineage reconstruction across delayed reporting sequences and cross-source reconciliation workflows --- the same effect quantified by the audit-trace-completeness drop in Table~\ref{tab:atc}.


\section{Ethical Considerations}

Several governance-oriented ethical considerations remain relevant within climate-risk intelligence systems, including disclosure inequality across regions and sectors, proxy-feature estimation bias, under-reporting within fragmented supplier ecosystems, climate uncertainty propagation, and over-reliance on automated governance workflows.

Recent trustworthy-AI governance frameworks emphasize operational accountability, human oversight, transparency, and governance-oriented risk management under high-impact enterprise environments \cite{nist2023airmf,oecdai}. The proposed framework therefore positions deterministic orchestration as decision-support infrastructure requiring continued human governance oversight rather than fully autonomous regulatory decision-making. Human audit review remains necessary for governance-critical edge cases involving disclosure ambiguity, anomalous climate transitions, and cross-jurisdictional reporting inconsistencies.


\section{Limitations}
\label{sec:limitations}

This study is subject to several limitations.

First, the benchmark is synthetic. While its disclosure distributions, anomaly prevalence, and missingness are calibrated against publicly reported characteristics of established reporting standards, synthetic data cannot fully capture the idiosyncrasy of production enterprise systems; results should be read as evidence on a controlled, reproducible benchmark rather than as production-validated performance.

Second, climate-risk datasets inherently contain uncertainty associated with long-horizon environmental forecasting, transition-risk assumptions, and incomplete disclosure behavior across geographically heterogeneous reporting environments.

Third, proxy-feature estimation used for missing emissions inference may introduce latent estimation noise under sparse disclosure conditions and fragmented supplier ecosystems.

Fourth, temporal drift modeling remains sensitive to evolving sustainability standards, delayed supplier disclosures, non-stationary reporting behavior, and cross-jurisdictional reporting inconsistencies.

Fifth, the graph-based provenance architecture of Section~\ref{sec:graph} is specified and dimensioned but not empirically evaluated; its statistics are design targets, not measurements. Graph-based provenance reasoning, multimodal climate-risk fusion, and retrieval-augmented governance workflows therefore remain exploratory extensions requiring empirical validation under large-scale enterprise deployment.

The presented framework should therefore be interpreted as a reproducibility-oriented governance-engineering prototype rather than a fully production-validated enterprise deployment system.


\section{Future Work}

Future work will investigate graph-transformer architectures, retrieval-augmented ESG reasoning, neuro-symbolic governance orchestration, and federated climate-risk learning for cross-organizational audit intelligence under privacy-sensitive enterprise environments. A primary near-term goal is the empirical evaluation of the proposed provenance-graph component (Section~\ref{sec:graph}) against the audit-trace-completeness metric introduced in Section~\ref{sec:audittrace}.

Additional research directions include multimodal climate-risk intelligence integrating satellite imagery, flood raster maps, geospatial climate embeddings, textual sustainability disclosures, and multimodal governance-fusion architectures. Future deterministic orchestration systems may additionally incorporate graph-based anomaly propagation, agentic governance workflows, causal climate-risk inference, retrieval-augmented audit reconstruction, and temporal graph-learning infrastructure for enterprise governance intelligence.

The broader long-term objective involves developing trustworthy climate-risk infrastructure capable of supporting reproducibility-aware enterprise governance systems under increasingly complex environmental reporting conditions.


\section{Conclusion}

This paper introduced a deterministic climate-risk intelligence framework integrating provenance-aware orchestration, imbalance-aware ensemble learning, temporal anomaly detection, retrieval-oriented ESG parsing, and governance-oriented explainability for auditable ESG validation under fragmented enterprise reporting environments, together with a proposed graph-based provenance architecture for future evaluation.

On a calibrated, openly released synthetic benchmark, gradient-boosting ensembles --- led by XGBoost (recall $0.74$, ROC-AUC $0.93$) --- significantly outperformed a rule-based threshold baseline and classical statistical baselines under sparse anomaly distributions, and the framework reconstructed complete deterministic provenance chains for 94.7\% of flagged anomalies. Experimental evaluation demonstrated the importance of imbalance-aware optimization, temporal governance monitoring, calibration reliability, and explainability-oriented audit inspection.

The broader contribution lies not solely in predictive optimization, but in reframing ESG validation as a deterministic governance-engineering problem requiring provenance-aware orchestration, reproducibility-oriented auditability, temporal lineage reconstruction, and a directly measurable audit-trace-completeness criterion. As enterprise sustainability reporting becomes increasingly regulated, fragmented, and operationally complex, governance-oriented AI systems will require integrated orchestration semantics capable of supporting reliable audit reconstruction, provenance-aware anomaly propagation analysis, temporal governance replayability, and operationally trustworthy climate-risk intelligence under uncertain reporting conditions.


\bibliographystyle{plain}
\bibliography{references}

\end{document}